\pdfoutput=1

\documentclass[11pt]{article}

\usepackage[final]{acl}

\usepackage{times}
\usepackage{latexsym}

\usepackage[T1]{fontenc}

\usepackage[utf8]{inputenc}

\usepackage{microtype}

\usepackage{inconsolata}

\usepackage{graphicx}

\usepackage{amsmath}
\usepackage{graphicx}
\usepackage{booktabs}
\usepackage{amssymb}
\usepackage{transparent}
\usepackage{multirow}
\usepackage{tikz}
\usepackage[capitalize]{cleveref}
\usepackage{pifont}
\usepackage[super]{nth}
\usepackage[shortlabels]{enumitem}
\usepackage{quoting}
\usepackage{subcaption}
\usepackage{adjustbox}
\usepackage{xcolor}
\usepackage{ulem}
\usepackage{scalerel}
\usepackage{dcolumn}
\usepackage{url}

\usepackage{array}
\usepackage{booktabs}
\setenumerate[1]{label=C\arabic*.}
\newcolumntype{d}[1]{D{.}{.}{#1}}

\usepackage[whole]{bxcjkjatype}
\usepackage{pxrubrica}

\crefformat{section}{\S#2#1#3} 
\crefformat{subsection}{\S#2#1#3}
\crefformat{subsubsection}{\S#2#1#3}

\title{Robust ASR Error Correction with Conservative Data Filtering}

\author{Takuma Udagawa, Masayuki Suzuki, 
Masayasu Muraoka,  Gakuto Kurata
 \\
IBM Research AI \\
\texttt{takuma.udagawa@ibm.com, \{szuk,mmuraoka,gakuto\}@jp.ibm.com}}

\begin{document}
\maketitle
\begin{abstract}
Error correction (EC) based on large language models is an emerging technology to enhance the performance of automatic speech recognition (ASR) systems.
Generally, training data for EC are collected by automatically pairing a large set of ASR hypotheses (as sources) and their gold references (as targets).
However, the quality of such pairs is not guaranteed, and we observed various types of noise which can make the EC models brittle, e.g. inducing overcorrection in out-of-domain (OOD) settings.
In this work, we propose two fundamental criteria that EC training data should satisfy: namely, EC targets should (1) improve linguistic acceptability over sources and (2) be inferable from the available context (e.g. source phonemes).
Through these criteria, we identify low-quality EC pairs and train the models not to make any correction in such cases, the process we refer to as conservative data filtering.
In our experiments, we focus on Japanese ASR using a strong Conformer-CTC as the baseline and finetune Japanese LLMs for EC.
Through our evaluation on a suite of 21 internal benchmarks, we demonstrate that our approach can significantly reduce overcorrection and improve both the accuracy and quality of ASR results in the challenging OOD settings.
\end{abstract}

\section{Introduction}
\label{sec:introduction}

\begin{table*}[t!]
\centering
\begin{adjustbox}{max width=0.78\textwidth}
\setlength\tabcolsep{6pt}
\begin{tabular}{lccc}
\hline
\multicolumn{2}{c}{} & ASR Hypothesis (Source $W^S$) & Gold Reference (Target $W^T$) \\
\hline
\hline\\[-1.1em]
\multicolumn{2}{c}{\multirow{3}{*}{Clean}} & に雑音を\underline{蒸}したもの & に雑音を\underline{付加}したもの \\
& & [ni zatsuon o \underline{fuka} shita mono] & [ni zatsuon o \underline{fuka} shita mono] \\
&  &  (\textit{to which noise is \underline{steamed}}) & (\textit{to which noise is \underline{added}}) \\
\hline\\[-1.1em]
\multirow{6}{*}{Noisy\:} & \multirow{3}{*}{\shortstack[c]{Incorrect/\\Unnecessary}} & しかし一対一\underline{の}場合ですと & しかし一対一場合ですと \\
& & [shikashi ittaiichi \underline{no} baai desuto] & [shikashi ittaiichi baai desuto] \\
& & (\textit{but in case \underline{of} one-to-one}) & (\textit{but in case one-to-one})  \\[-0.1em]
\cmidrule{2-4}\\[-1.4em]
& \multirow{3}{*}{Uninferable} & 男の人\underline{はぐらい}ですかね & 男の人\underline{の方がいい}ですかね \\
& & [otokonohito \underline{wa gurai} desukane] & [otokonohito \underline{noho:ga ii} desukane] \\
& & (\textit{would a male person be \underline{about}}) & (\textit{would a male person be \underline{better}}) \\
\hline
\end{tabular}
\end{adjustbox}
\caption{\label{tab:ec_examples}
Clean and noisy examples observed in our Japanese EC training data. Phonemes are shown in square brackets [] and English translation in round brackets (). Targets can be naturally inferred from the erroneous sources in the clean cases, while incorrect, unnecessary, or uninferable corrections are required in the noisy cases.}
\end{table*}

\begin{figure*}[t!]
    \includegraphics[width=0.99\textwidth]{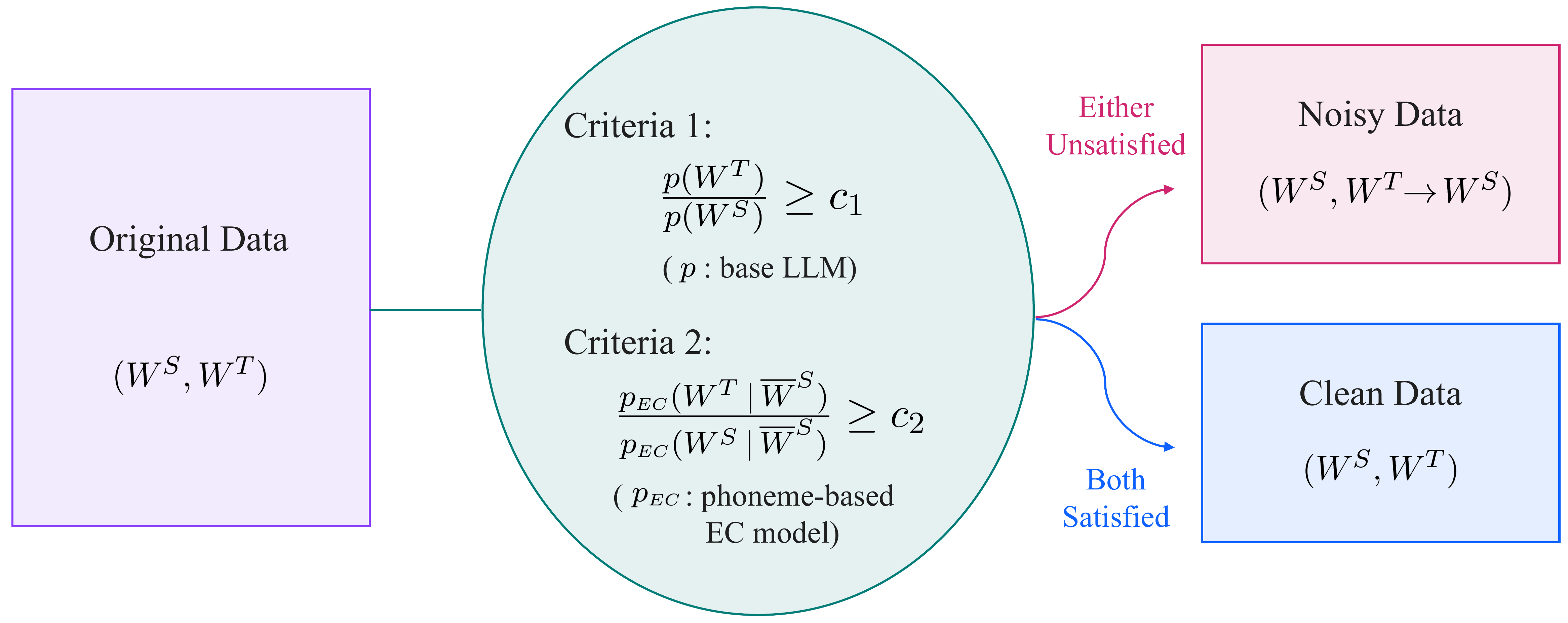}
    \caption{An illustration of our conservative data filtering. Precise details and terminologies are explained in \cref{sec:methods}.
    }
    \label{fig:conservative_filtering}
\end{figure*}

Automatic speech recognition (ASR) is the task of transcribing human speech into readable text, which is of practical use in various applications. 
In contrast to the traditional hybrid approach \citep{sak2014long}, modern ASR systems are trained in an end-to-end manner using a large parallel corpus of acoustic speech paired with gold transcriptions \citep{prabhavalkar2017comparison,li2022recent}.
Despite their huge success, end-to-end ASR systems have limited linguistic knowledge due to the difficulty of leveraging unpaired text-only data which exist in abundance \citep{penedo2023refinedweb}. 

Error correction (EC) is an effective strategy to correct linguistic errors produced by such ASR systems \citep{errattahi2018automatic,guo2019spelling}.
Recently, large language models (LLMs) pretrained on massive text-only data have shown promising results for this purpose \citep{ma2023nbest,chen2024hyporadise}.
While several works explore the zero-shot or in-context learning capability of LLMs \citep{ma2023can,yang2023generative}, finetuning LLMs with sufficient EC training data remains critical to impart the knowledge of ASR-specific error patterns and desired corrections \citep{mani2020asr,leng2021fastcorrect,radhakrishnan-etal-2023-whispering,wang2023hypr,chen2024its}

Generally, training data for EC are collected by automatically pairing the ASR hypothesis (source) and its gold transcription (target), and the task is formulated as sequence transduction from the source to target \citep{guo2019spelling}.
However, the quality of such pairs is not guaranteed: in fact, we observed various types of noise which require incorrect, unnecessary, or uninferable corrections that are unreasonable to be predicted from the source.
We show some illustrative examples in Table \ref{tab:ec_examples}.

Training EC models on such noisy data can amplify \textit{overcorrection}, which is a typical problem in current EC \citep{ma2023can,leng2023softcorrect}.
However, existing works largely overlook the existence of such noise and apply minimal data filtering, e.g. simply discard pairs with large edit distance \citep{hrinchuk2020correction,Zhao2021BARTBS}.

In this study, we propose two fundamental criteria that EC training data should satisfy in general. Specifically, we ensure that EC targets
\begin{enumerate}[topsep=2pt, itemsep=0pt, leftmargin=.3in, parsep=2pt]
  \item improve linguistic acceptability over sources
  \item are inferable from the available context (e.g. source phonemes)
\end{enumerate}
Based on these criteria, we identify low-quality EC pairs and train the models to avoid making any correction on them. 
Such conservative behavior is often crucial to reduce overcorrection and improve robustness, esp. in the out-of-domain (OOD) settings \citep{li2024crossmodal}.
The overall flow of our data filtering strategy is shown in Figure \ref{fig:conservative_filtering}.

In our experiments, we focus on Japanese ASR using an internal Conformer-CTC as the baseline \citep{lee2021intermediate}. 
For EC, we finetune opensource Japanese LLMs, namely Swallow-Mistral 7B\footnote{\url{https://huggingface.co/tokyotech-llm/Swallow-MS-7b-v0.1}} and Sarashina-2 7B\footnote{\url{https://huggingface.co/sbintuitions/sarashina2-7b}}, and evaluate the performance on 21 internal benchmarks comprised of various domains.
Through our experiments, we confirm that our approach can significantly reduce overcorrection and robustly improve ASR results in the most challenging OOD settings.

\section{Related Work}
\label{sec:related_work}

In the existing literature, EC primarily focuses on the \textit{in-domain} setup where models are trained and evaluated over the same domain \citep{guo2019spelling,mani2020asr,wang2020asr,leng2021fastcorrect,ma2023nbest}. 
Recently, \citet{li2024crossmodal} proposed a \textit{low-resource OOD} setup where EC models are finetuned on a limited amount of target domain data to generalize beyond in-domain data.
However, target domains of EC are conceptually broad or even open-ended, so it is desirable that EC models work reliably in any target domain without prior knowledge or finetuning.
In this study, we focus on the most challenging \textit{zero-resource OOD} setup to develop general-purpose EC models which work out of the box in a variety of domains.

Despite the recent progress, overcorrection remains a major challenge in EC, esp. in the OOD setup.
To alleviate this issue, constrained decoding \citep{Zhao2021BARTBS,ma2023nbest,ma2023can} restricts or biases the correction towards retaining the original ASR hypotheses.
\citet{li2024crossmodal} use a representative data source and partially train the models to copy the input to induce conservative behavior.
In complementary to their approach, we focus on the \textit{quality} of EC data and apply sophisticated data filtering, which is a novel aspect of our approach that works much more effectively than existing filtering based on edit distance \citep{hrinchuk2020correction,Zhao2021BARTBS,ma2023nbest}.

Typically, ASR errors originate from confusing phonetically similar words and phrases.
Therefore, supplementing EC models with phonetic/acoustic information can help improve their performance \citep{wang2020asr,dutta2022error,higuchi2023harnessing,li2024crossmodal}.
In this study, we use the source phonemes as an additional input, which can be easily handled by the text-based LLMs.
When available, the full N-best hypotheses can be used as input to provide richer clues on where the ASR systems are confused \citep{zhu2021improving,ma2023nbest}.
However, for both simplicity and computational efficiency, we only use the  1-best hypothesis (i.e. top ASR prediction) in our experiments.

\section{Methods}
\label{sec:methods}

EC can be formulated as a sequence transduction task from the ASR hypothesis (source) to the gold reference (target).
Formally, let $(W^S, W^T)$ denote the source and target sequence pair.
In the simplest setting, the EC model is trained to estimate $p_{\scriptscriptstyle EC} (W^T \,|\, W^S)$ with the expectation of transforming an error-prone source into a clean target.

In this study, we also incorporate the source phonemes $\overline{W}^S$ as an additional input, in which case the EC model estimates $p_{\scriptscriptstyle EC} (W^T \,|\, W^S, \overline{W}^S)$.
Source phonemes are obtained from our ASR system (\cref{sec:experimental_setup}) and represented in hiragana, one of the Japanese syllabaries, which can be easily consumed by Japanese LLMs. Below is an example: 
\begin{itemize}[topsep=2pt, itemsep=0pt, leftmargin=.2in, parsep=2pt]
    \item $W^S$: {\normalsize また\underline{海}属性に関しては} \\
    \phantom{ $W^S$: }[mata \underline{kai} zokuse: ni kanshitewa] \\
    \phantom{ $W^S$: }(\textit{also in terms of \underline{sea} attribute})
    \item $\overline{W}^S$: {\normalsize また\:\underline{かい}\:ぞくせー\:に\:かんしてわ} 
    \item $W^T$: {\normalsize また\underline{下位}属性に関しては} \\
    \phantom{ $W^S$: }[mata \underline{kai} zokuse: ni kanshitewa] \\
    \phantom{ $W^S$: }(\textit{also in terms of \underline{subordinate} attribute})
\end{itemize}

Generally, training data of EC can be collected at scale by automatically pairing the hypotheses and gold references in the ASR system's training data.\footnote{Although the ASR systems are directly trained on these datasets, they usually make sufficient errors for EC models to learn from. One can virtually increase the amount of errors through noise injection \citep{Zhao2021BARTBS} or data partitioning to avoid training on each partition \citep{hrinchuk2020correction}.
}
However, not all source-target pairs are suitable for training EC models, as we observed various types of noise (illustrated in Table \ref{tab:ec_examples}).
To address this issue, we propose two fundamental criteria that high-quality EC pairs should satisfy.


\paragraph{Criteria 1: EC targets should improve linguistic acceptability over sources.} 
The main objective of EC is to resolve linguistic errors in the ASR system's predictions and improve linguistic acceptability.
While the gold reference usually contains cleaner text, this is not always the case, e.g. due to speaker disfluency in spontaneous speech or noisy transcriptions. 
In addition, Japanese is a language with rich orthographic variation where multiple valid spellings exist \citep{ohsugi2022japanese,karita2023lenient}.
For instance, the correction is not necessary if the source transcribes a \textit{bottle} as {\normalsize 瓶} [bin] while the target transcribes as {\normalsize ビン} [bin], since both spellings are equally acceptable.

To improve robustness, EC models should only focus on apparent mistakes and resolve them accurately.
One simple way to express this criteria is based on the following equation:
\begin{equation}
\frac{p(W^T)}{p(W^S)} \geq c_1
\label{eq:c1_filter}
\end{equation}
Here, $p(W^S)$ and $p(W^T)$ denote the likelihoods of the source and target, which can be computed using any language model. In this study, we simply use the base Japanese LLM.
$c_1$ denotes the threshold, set to $1$ by default, which can control the strength of the filter.
Intuitively, $(W^S, W^T)$ that do not satisfy this criteria do not sufficiently improve the linguistic acceptability, indicating the correction is incorrect or unnecessary.

\paragraph{Criteria 2: EC targets should be inferable from the available context.}
Existing works assume that EC targets are generally inferable from the source.
However, this is not always the case: in fact, expert evaluation revealed that about one-third of the errors cannot be corrected from the source alone \citep{Zhao2021BARTBS}.
This is mainly attributed to the large phonetic discrepancy between the source and target, e.g. caused by environmental noise or incapability of the ASR system.

A robust EC model should only make the correction when it is inferable from the available context.
To express this criteria, we quantify the degree of inferability from the source phonemes using the following equation:
\begin{equation}
\frac{p_{\scaleto{EC}{4pt}}(W^T \,|\, \overline{W}^S)}{p_{\scaleto{EC}{4pt}}(W^S \,|\, \overline{W}^S)} \geq c_2
\label{eq:c2_filter}
\end{equation}
Here, $p_{\scriptscriptstyle EC}$ is a baseline EC model trained only using the source phonemes as input.
In this study, we finetune the base Japanese LLM following the procedure described in \cref{sec:experimental_setup}.
Again, $c_2$ denotes the threshold which can be set to $1$ by default.
Intuitively, $(W^S, W^T)$ that do not satisfy this criteria cannot be easily inferred from the available context, namely source phonemes in our case.

It is worth noting that edit distance is not a suitable measure of inferability. For instance, the uninferable example in Table \ref{tab:ec_examples} has a relatively small edit distance but is very difficult to be inferred.
In contrast, the following example is quite dissimilar in terms of edit distance but can be more naturally inferred from the source phonemes:
\begin{itemize}[topsep=2pt, itemsep=0pt, leftmargin=.2in, parsep=2pt]
    \item $W^S$: {\normalsize \underline{そうか検出}で} [\underline{so:ka kensyutsu} de] \\
    \phantom{ $W^S$: }(\textit{based on the \underline{I see detection}})
    \item $W^T$: {\normalsize \underline{相関係数}で} [\underline{so:kan ke:suu} de] \\
    \phantom{ $W^T$: }(\textit{based on the \underline{correlation coefficient}})
\end{itemize}
\noindent

\bigskip

\begin{figure}[t!]
    \includegraphics[width=0.99\columnwidth]{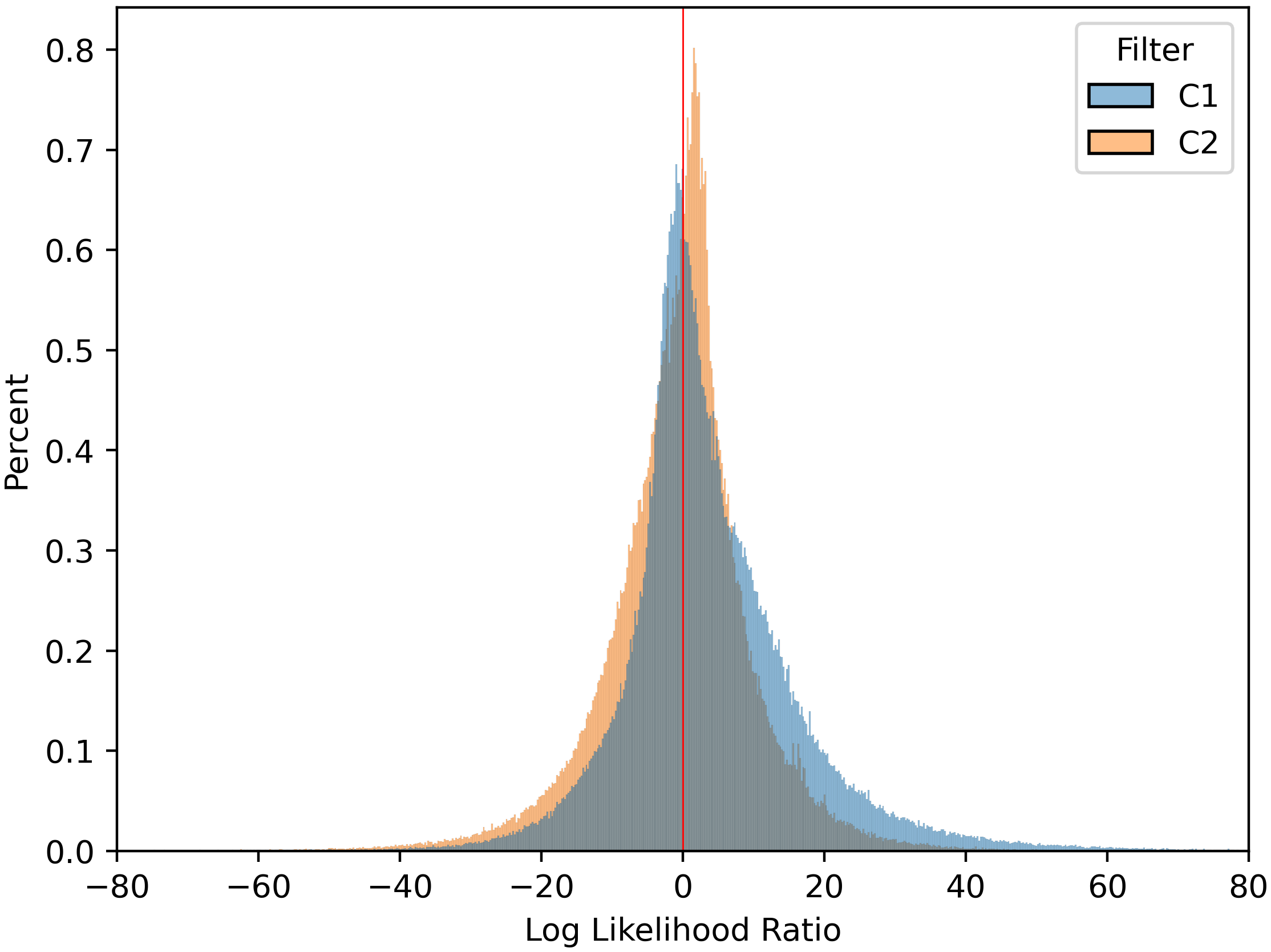}
    \caption{Log-likelihood ratios for the two criteria, i.e. $\log \frac{p(W^T)}{p(W^S)}$ for C1 and $\log \frac{p_{\scaleto{EC}{2.5pt}}(W^T \,|\, \overline{W}^S)}{p_{\scaleto{EC}{2.5pt}}(W^S \,|\, \overline{W}^S)}$ for C2. Red line shows the default threshold ($c_1 = c_2 = 1$).
    }
    \label{fig:log_likelihood_ratio}
\end{figure}

Based on the above criteria (C1 and C2), we identify low-quality EC pairs and train the models to avoid making any correction on them by simply replacing the target with source ($W^T \rightarrow W^S$): see Figure \ref{fig:conservative_filtering} for an illustration.
We found this approach more effective than discarding the noisy pairs, since the model is explicitly trained to be conservative on noisy or otherwise ambiguous examples. 

Note that both criteria are defined based on the likelihood ratio between the source and target (eq. \ref{eq:c1_filter}, \ref{eq:c2_filter}).
In Figure \ref{fig:log_likelihood_ratio}, we plot the distribution of the (log-)likelihood ratio for each criteria in our training data, using Swallow-Mistral 7B as the LLM.\footnote{Statistics based on Sarashina-2 7B are provided in Appendix \ref{sec:sarashina2_7b}, where we observed similar results.}
We can verify that a non-negligible portion of the pairs do not satisfy the criteria, suggesting noisy pairs are prevalent in EC training data.
For additional examples of the filtered/non-filtered pairs, we refer the reader to \cref{sec:filtered_data_examples}.

Out of the whole training data, our ASR baseline predicts the exact gold reference (i.e. ${W}^S = {W}^T$) in about 34\% of the cases.
Therefore, the EC model effectively learns to make a correction (i.e. ${W}^S \neq {W}^T$) in only 66\% of the cases.
Of these effective pairs, 34\% are classified as noisy based on our C1 filter, 33\% based on C2 filter, and 42\% when combined.
While this results in even fewer examples to learn from, we can expect the model to focus on clearer errors and improve OOD robustness.
Our approach is also in line with the principle that data quality can be more important than quantity for LLM alignment \citep{zhou2023lima}.

\section{Experimental Setup}
\label{sec:experimental_setup}

\begin{table*}[t!]
\centering
\begin{adjustbox}{max width=0.99\textwidth}
\setlength\tabcolsep{4pt}
\begin{tabular}{c||c||ccc|ccc|ccc|ccc|ccc}
\hline
\multirow{2}{*}{Test} & \underline{Orig.} & \multicolumn{3}{c|}{\underline{No Filter}} & \multicolumn{3}{c|}{\underline{C1 Only}} & \multicolumn{3}{c|}{\underline{C2 Only}} & \multicolumn{3}{c|}{\underline{C1+C2}} & \multicolumn{3}{c}{\underline{Inv. C1+C2}} \\
 & CER & CER & \%EC & \%LA & CER & \%EC & \%LA & CER & \%EC & \%LA & CER & \%EC & \%LA & CER & \%EC & \%LA \\
\hline
\hline
1& \textbf{\phantom{0}6.66} & \phantom{0}9.28 & 38.5 & 62.7 & \phantom{0}7.67 & 14.9 & 57.7 & \phantom{0}8.14 & 21.8 & 55.3 & \phantom{0}7.97 & 16.7 & 69.0 & \phantom{0}8.01 & \phantom{0}9.2 & 43.8 \\
2& \phantom{0}8.18 & \phantom{0}7.65 & 55.4 & 60.6 & \textbf{\phantom{0}7.13} & 32.6 & 71.7 & \phantom{0}7.42 & 27.8 & 63.9 & \phantom{0}7.44 & 21.4 & 72.6 & \phantom{0}8.49 & 11.0 & 35.8 \\
3& 20.66 & 21.55 & 42.7 & 48.1 & 20.46 & 20.5 & 59.9 & 20.65 & 21.8 & 49.7 & \textbf{20.15} & 13.1 & 60.9 & 20.85 & \phantom{0}7.6 & 46.4 \\
4& \textbf{18.74} & 21.18 & 26.1 & 57.1 & 19.56 & 13.8 & 66.7 & 20.10 & 12.9 & 55.6 & 19.17 & \phantom{0}7.5 & 57.7 & 20.78 & \phantom{0}8.6 & 50.0 \\
5& \phantom{0}6.13 & \phantom{0}7.50 & 25.0 & 75.4 & \phantom{0}7.04 & 16.4 & 87.3 & \phantom{0}7.05 & 18.8 & 81.8 & \phantom{0}7.08 & 14.3 & 87.0 & \textbf{\phantom{0}5.96} & \phantom{0}3.6 & 33.3 \\
6& \phantom{0}7.20 & \textbf{\phantom{0}6.89} & 15.4 & 50.0 & \textbf{\phantom{0}6.89} & 12.8 & 60.0 & \phantom{0}7.10 & \phantom{0}2.6 & \!\!100.0 & \textbf{\phantom{0}6.89} & \phantom{0}5.1 & 50.0 & \phantom{0}7.20 & \phantom{0}0.0 & - \\
7& \textbf{12.50} & 14.26 & 57.6 & 55.7 & 13.38 & 31.9 & 62.8 & 12.92 & 21.3 & 57.1 & 12.81 & 14.8 & 62.4 & 12.89 & 13.4 & 38.3 \\
8& \phantom{0}8.53 & \phantom{0}8.67 & 49.8 & 57.1 & \textbf{\phantom{0}8.39} & 26.0 & 64.2 & \phantom{0}8.54 & 23.8 & 58.8 & \phantom{0}8.45 & 14.7 & 68.3 & \phantom{0}8.62 & 11.5 & 44.7 \\
9& \phantom{0}8.47 & \phantom{0}7.82 & 35.7 & 51.5 & \textbf{\phantom{0}6.91} & 22.6 & 63.4 & \phantom{0}7.16 & 23.3 & 53.4 & \phantom{0}7.16 & 16.4 & 66.8 & \phantom{0}7.98 & 10.7 & 49.1 \\
10& \phantom{0}8.45 & \phantom{0}8.06 & 29.7 & 62.6 & \textbf{\phantom{0}7.34} & 17.4 & 67.0 & \phantom{0}7.86 & 14.9 & 64.1 & \phantom{0}7.67 & 12.2 & 65.0 & \phantom{0}8.52 & \phantom{0}4.8 & 51.9 \\
11& 19.77 & 21.00 & 47.4 & 59.3 & 22.41 & 22.0 & 67.5 & 19.89 & 19.8 & 58.3 & \textbf{19.70} & 14.3 & 59.6 & 20.35 & 11.6 & 52.4 \\
12& 12.02 & 12.08 & 46.8 & 59.0 & \textbf{11.34} & 23.3 & 59.6 & 11.72 & 20.9 & 51.0 & 12.24 & 11.4 & 64.3 & 12.40 & 17.4 & 43.5 \\
13& 13.06 & \textbf{12.64} & 31.9 & 53.0 & 12.83 & 10.9 & 61.8 & 12.83 & \phantom{0}9.6 & 70.0 & 12.95 & \phantom{0}5.4 & 35.3 & 12.91 & \phantom{0}3.5 & 54.5 \\
14& \textbf{26.10} & 27.88 & 48.5 & 54.9 & 26.86 & 18.9 & 65.9 & 26.81 & 14.6 & 50.0 & 26.79 & 11.6 & 81.5 & 26.45 & 13.7 & 40.6 \\
15& 15.23 & 16.36 & 47.4 & 51.4 & 15.34 & 22.5 & 64.2 & 15.47 & 17.8 & 53.7 & \textbf{15.20} & 12.2 & 55.4 & 15.88 & 13.9 & 43.2 \\
16& 12.03 & 14.08 & 54.7 & 62.2 & 11.36 & 22.7 & 73.5 & \textbf{10.74} & 24.7 & 75.7 & 11.65 & 16.7 & 64.0 & 12.32 & \phantom{0}2.0 & \phantom{0}0.0 \\
17& \phantom{0}9.98 & \phantom{0}9.53 & 28.0 & 61.6 & \phantom{0}9.35 & 18.2 & 66.5 & \phantom{0}9.52 & 15.5 & 62.6 & \textbf{\phantom{0}9.31} & 12.5 & 65.7 & \phantom{0}9.92 & \phantom{0}4.6 & 56.2 \\
18& 14.52 & 16.81 & 56.0 & 50.0 & 15.70 & 34.0 & 52.9 & 14.28 & 22.0 & 45.5 & \textbf{14.03} & 10.0 & 60.0 & 15.02 & \phantom{0}8.0 & 50.0 \\
19& \phantom{0}6.89 & \phantom{0}5.84 & 56.0 & 75.0 & \phantom{0}5.76 & 40.0 & 85.0 & \phantom{0}6.17 & 34.0 & 82.3 & \textbf{\phantom{0}5.69} & 22.0 & 90.9 & \phantom{0}6.66 & \phantom{0}4.0 & \!\!100.0 \\
20& \phantom{0}7.81 & \phantom{0}7.45 & 69.6 & 73.4 & \textbf{\phantom{0}7.33} & 32.5 & 83.7 & \phantom{0}7.61 & 22.0 & 81.9 & \phantom{0}7.66 & 27.8 & 86.7 & \phantom{0}7.83 & \phantom{0}5.0 & 52.6 \\
21& \phantom{0}5.76 & \phantom{0}6.20 & 41.4 & 53.1 & \phantom{0}5.91 & 20.6 & 58.2 & \phantom{0}5.76 & 20.6 & 56.3 & \textbf{\phantom{0}5.39} & 11.6 & 67.2 & \phantom{0}5.64 & \phantom{0}7.2 & 52.8 \\
\hline
\hline
Avg. & 11.84 & 12.51 & 43.0 & 58.8 & 11.86 & 22.6 & 66.7 & 11.80 & 19.5 & 63.2 & \textbf{11.69} & 13.9 & 66.2 & 12.13 & \phantom{0}8.2 & 47.0 \\
\hline
$<$Orig. & - & 38.1 & - & - & 57.1 & - & - & 57.1 &- & -& \textbf{71.4} &- &- & 28.6 & -& -\\
\hline
\hline
\end{tabular}
\end{adjustbox}
\caption{\label{tab:swallow_mistral_results}
Results of EC using Swallow-Mistral 7B. Based on 21 internal test sets, we compute the macro average score for each metric (Avg.) and the ratio of test sets where CER is improved over the original ASR ($<$Orig.).
}
\end{table*}

\paragraph{ASR System} For the ASR baseline, we use an internal Conformer-CTC developed for commercial use cases. 
The acoustic model is a CTC \citep{graves2006connectionist} with 240-dimensional logmel-derived features every 40 milliseconds as input, consisting of 10 conformer layers \citep{Gulati2020Conformer}, followed by an output layer of 42 Japanese phonemes including the blank symbol.
For inference, a static graph for graph decoding is created using a word n-gram model and a dictionary representing the mapping between words ($W^S$) and their phoneme sequences ($\overline{W}^S$).
In total, our training data consists of 8000 hours of transcribed speech with little or no overlap between our benchmarking domains.

\paragraph{EC Model} For EC, we finetune two Japanese LLMs, namely Swallow-Mistral 7B \citep{fujii2024continual,okazaki2024building} and a more recent Sarashina-2 7B, on a subset of the ASR training data ensuring a 1:1 mixture of read and spontaneous speech.
For finetuning, we use LoRA \citep{hu2021lora} with rank $r=32$ and scaling factor $\alpha=16$.
The effective batch size is set to 128 on a single A100 GPU, and the learning rate is 5e--4 annealed with a cosine scheduler.
All EC models are trained for a total of 1000 steps, since we observed more steps led to overfitting in the OOD setup.
For inference, we use greedy decoding, which we found to be efficient yet effective.

For C2 filtering, we train the phoneme-based EC model to predict the target $W^T$ only using the source phonemes $\overline{W}^S$ as input.
Otherwise, models are trained with both the source phonemes $\overline{W}^S$ and the source hypothesis $W^S$ as input.

As an ablation study, we compare the performance of EC models without any data filtering (No Filter), with C1 and C2 filtering applied independently (C1/C2 Only), and with both filtering applied in combination (C1+C2).
In addition, to confirm that noisy pairs are less effective for EC training, we also experiment with an inverse filtering of C1+C2, considering the noisy pairs as clean and vice versa (Inv. C1+C2).

\paragraph{Evaluation} We evaluate EC performance on 21 internal benchmarks comprised of various domains.
Details of each benchmark are provided in Table \ref{tab:benchmark_info}.
All EC models are evaluated in the zero-resource OOD setup without any domain adaptation.

As for the evaluation metrics, we primarily focus on character error rate (\textbf{CER$^\downarrow$}) which is standardly used for ASR.
To quantify the degree of overcorrection, we also measure the percentage of source hypotheses altered after EC (\textbf{\%EC$^\downarrow$}).
Finally, we measure the percentage of hypotheses where the linguistic acceptability is improved after EC (\textbf{\%LA$^\uparrow$}).
To measure \%LA, we compare the masked language modeling score \citep{salazar-etal-2020-masked} of the hypothesis before and after EC using Japanese DeBERTa V2 large\footnote{\url{https://huggingface.co/ku-nlp/deberta-v2-large-japanese}}.
While DeBERTa is relatively small compared to recent LLMs, it can take into account the full (bidirectional) context of the hypothesis and effectively assess its linguistic acceptability \citep{udagawa2022effect}.

\section{Results and Discussion}
\label{sec:experimental_results}

In Table \ref{tab:swallow_mistral_results}, we report the results of our experiments using Swallow-Mistral 7B as the Japanese LLM.
Results based on the recent Sarashina-2 7B are provided in Appendix \ref{sec:sarashina2_7b}, where we observed similar trends with even better performance.

Focusing on Swallow-Mistral 7B, there is no single approach which outperforms all others due to the diversity of the test sets.
However, we can still draw several conclusions from the overall metrics.
First, compared to the original ASR results, we can verify that EC without data filtering drastically worsens CER on average (11.84 $\rightarrow$ 12.51).
This is mainly attributed to the overcorrection problem, as we can see a large portion of the hypotheses (43.0\% on average) are altered by EC. 
Such aggressive behavior can be helpful in some occasions (e.g. Test 13) but generally too risky in the OOD setup, leading to modest or even severe performance degradation (e.g. in Test 1, 4, 16, 18, to count a few).

In contrast, by applying our C1 filtering, we can substantially alleviate the degradation of CER (11.84 $\rightarrow$ 11.86) by almost halving the frequency of corrections (22.1\% on average).
This shows that EC can be kept more accurate and conservative by training on cleaner pairs which improve linguistic acceptability.
Our C2 filtering also has a similar benefit and makes the EC model more robust in the OOD setup, outperforming the original ASR results in 57.1\% (12/21) of the test cases.

In addition, by combining our C1+C2 filtering, we can further cut down the frequency of corrections to 13.9\% on average.
Through this conservative behavior, we could significantly improve the OOD robustness of EC and reduce the original CER in 71.4\% (15/21) of the test sets.
This result demonstrates that both C1 and C2 filters help EC focus on clear and fixable ASR errors whilst ignoring more controversial ones.

To verify that clean (rather than noisy) portions of the data contribute to this improvement, we also experimented with the inverse filtering of C1+C2. 
Generally, we confirmed that inverse filtering worsens CER on average (11.84 $\rightarrow$ 12.13) and only improves upon the original ASR in 28.6\% (6/21) of the test sets.
Therefore, noisy pairs are much less effective for accurate EC.
While the frequency of correction is drastically suppressed (8.2\% on average), this is largely attributed to the difficulty of learning from noisy examples and overlooking clear errors.
In a few cases (e.g. Test 5), we found inverse filtering to be quite competitive, which suggests that noisy pairs still include useful examples for some domains.
We expect that our filtering can be improved for such domains by appropriately tuning the thresholds (e.g. lowering $c_1$ and $c_2$) to include useful pairs of borderline quality.

Finally, in terms of the linguistic acceptability (\%LA), we generally see improvement through EC: this indicates that EC is at least successful in resolving linguistic errors and improving ASR quality, even if by deviating from the ground truth \citep{Zhao2021BARTBS}.
Naturally, our C1 filtering consistently strengthens this desirable property by explicitly taking this criteria into account (eq. \ref{eq:c1_filter}).


\begin{table}[t!]
\centering
\begin{adjustbox}{max width=0.95\columnwidth}
\setlength\tabcolsep{4pt}
\begin{tabular}{c||ccc|ccc}
\hline
\multirow{2}{*}{Test} & \multicolumn{3}{c|}{\underline{Edit Dist. (0.5)}} & \multicolumn{3}{c}{\underline{Edit Dist. (0.25)}}  \\
 & CER & \%EC & \%LA & CER & \%EC & \%LA  \\
\hline
\hline
Avg. & 12.65 & 42.5 & 59.3 & 12.85 & 42.6 & 57.9 \\
\hline
$<$Orig. & 42.9 & - & - & 33.3 & - & - \\
\hline
\hline
\end{tabular}
\end{adjustbox}
\caption{\label{tab:edit_results_swallow}
Results of EC using Swallow-Mistral 7B with data filtering based on maximum edit distance. 
}
\end{table}

\bigskip

As an additional experiment, we also evaluated the results of EC with data filtering based on maximum edit distance \citep{hrinchuk2020correction,Zhao2021BARTBS,ma2023nbest}.
In this approach, EC pairs with normalized edit distance above a certain threshold are simply discarded from the training data.\footnote{Before computing edit distance, we normalized source and target texts by converting them into hiragana using \textit{pykakasi}: \url{https://github.com/miurahr/pykakasi}.}
We chose the commonly used thresholds of 0.5 and 0.25, which discard 1\% and 5\% of the whole training data, respectively.

The results are shown in Table \ref{tab:edit_results_swallow}.
We can confirm that filtering based on edit distance fails to improve CER and hardly reduces \%EC.
This demonstrates that such simple filtering is insufficient to improve the robustness of EC in the challenging OOD setup, regardless of its widespread usage.

Finally, to verify that our claims hold for a different Japanese LLM, we also experimented using Sarashina-2 7B. 
As discussed in Appendix \ref{sec:sarashina2_7b}, we can draw similar conclusions with even better performance, achieving an average CER of 11.41 in the best case and outperforming the original ASR results in 85.7\% (18/21) of the test sets.
Therefore, our approach is generalizable using other LLMs and we can expect to further improve performance by leveraging more powerful LLMs.

\section{Conclusion}
\label{sec:conclusion}

EC is an emerging technology to boost the performance of ASR by harnessing the power of LLMs.
However, current EC remains brittle, often degrading performance due to overcorrection in the OOD setup, which hinders its practical application.

In this study, we first focused on the quality of EC training data and proposed a method to identify noisy data based on two fundamental criteria.
Second, we revealed that EC data contains a considerable proportion of such noisy pairs, which can be effectively handled through our conservative data filtering.
Finally, we demonstrated that our approach can significantly alleviate the overcorrection problem and improve the robustness of EC in the challenging zero-resource OOD setup.

In contrast to the existing filtering methods (e.g. based on edit distance), we expect the quality of our data filtering to keep improving as the underlying LLMs become more powerful and accurate, which is a notable trend in the current literature.
In future work, we also plan to control for other important factors of data quality, such as diversity and representativeness \citep{suzuki2023extracting,yang2023representative}, to further improve the robustness of EC.
Overall, we expect our approach to be a foundational step towards developing general-purpose EC models applicable in any domain of interest, facilitating the utilization of LLM technology in the real-world scenarios.

\bibliography{custom}

\appendix

\section{Additional Data Examples}
\label{sec:filtered_data_examples}

\begin{table*}[t!]
\centering
\begin{adjustbox}{max width=0.86\textwidth}
\setlength\tabcolsep{6pt}
\begin{tabular}{cccc}
\hline\\[-1.1em]
\multirow{2}{*}{ASR Hypothesis (Source $W^S$)} & \multirow{2}{*}{Gold Reference (Target $W^T$)} & \multicolumn{2}{c}{Log-likelihood Ratios} \\
\cmidrule{3-4}
 & & C1 & C2 \\
\hline
\hline\\[-1.1em]
\underline{被験者人図}を表しています & \underline{被験者の人数}を表わしています & \multirow{3}{*}{$1.159$} & \multirow{3}{*}{$0.812$} \\\relax
 [\underline{hikensyaninzu} o arawashiteimasu] & [\underline{hikensya no ninzu:} o arawashiteimasu] & & \\
(\textit{it shows the \underline{human subject diagram}}) & (\textit{it shows the \underline{number of human subjects}}) & & \\[0.2em]
\hline\\[-1.1em]
で\underline{高校右下}ですね & で\underline{こうこういうモデル}です & \multirow{3}{*}{$0.680$} & \multirow{3}{*}{$-0.407$} \\\relax
 [de \underline{ko:ko: umoto} desune] & [de \underline{ko: ko:yu: moderu} desune] & & \\
(\textit{and \underline{high school lower right}}) & (\textit{and \underline{it's a model like this- this}}) & & \\[0.2em]
\hline\\[-1.1em]
\underline{Ａ二}の抽出方法ですが & \underline{二}の抽出方法ですが & \multirow{3}{*}{$-0.511$} & \multirow{3}{*}{$0.174$} \\\relax
 [\underline{e:ni} no chu:syutsu ho:ho: desuga] & [\underline{ni} no chu:syutsu ho:ho: desuga] & & \\
(\textit{in terms of the extraction method of \underline{A2}}) & (\textit{in terms of the extraction method of \underline{2}}) & & \\[0.1em]
\hline\\[-1.1em]
\underline{暖かく}なってまいりましてですね & \underline{たか高く}なってまいりましてですね & \multirow{3}{*}{$-1.187$} & \multirow{3}{*}{$-2.223$} \\\relax
 [\underline{atatakaku} natte mairimashite desune] & [\underline{takatakaku} natte mairimashite desune] & & \\
(it is getting \underline{warmer}) & (it is getting \underline{high- higher}) & & \\[0.1em]
\hline
\end{tabular}
\end{adjustbox}
\caption{\label{tab:filtered_data_examples}
Additional examples from the training data, along with their log-likelihood ratios for the two criteria: i.e. $\log \frac{p(W^T)}{p(W^S)}$ for C1 and $\log \frac{p_{\scaleto{EC}{2.5pt}}(W^T \,|\, \overline{W}^S)}{p_{\scaleto{EC}{2.5pt}}(W^S \,|\, \overline{W}^S)}$ for C2.
Based on the default thresholds ($c_1 = c_2 = 1$), both ratios must be above 0 to be considered clean (cf. \cref{sec:methods} for further details).
}
\end{table*}

In Table \ref{tab:filtered_data_examples}, we show additional examples of the EC pairs filtered/non-filtered based on our criteria.

In the first example, the uncommon noun {\normalsize 被験者人図} (\textit{human subject diagram}) is a transcription error and corrected into a more natural, similar-sounding phrase {\normalsize 被験者の人数} (\textit{number of human subjects}).
This is a perfectly valid example of EC and consequently assigned high log-likelihood ratios based on both criteria.

In the second example, the source (ASR hypothesis) is very unnatural and almost incomprehensible, significantly deviating from the target (gold reference).
Therefore, while the target is more natural and acceptable, there is no sufficient context to make it inferable and the pair is reasonably rejected based on the second criteria (C2).

In the third example, the beginning of the source {\normalsize Ａ} [e:] is a filler in the speech and not included in the target.
Such insertion errors are quite common in Conformer-CTC and hence regarded as inferable based on the baseline EC used in C2.
However, this is not a linguistic error in a genuine sense and is properly regarded as an unnecessary correction based on the first criteria (C1).

In the last example, the source is a perfectly valid sentence and even more natural than the target with speaker disfluency: {\normalsize たか高く} (\textit{high- higher}).
Therefore, it is unreasonable to expect an EC model to make such a correction, which can be safely ignored based on both criteria C1 and C2.

\section{Benchmark Details}
\label{sec:ood_benchmarks}

In Table \ref{tab:benchmark_info}, we provide a brief description of the benchmarks used in our experiments.
To evaluate ASR from multiple aspects, our test sets encompass a wide range of domains with various difficulties and characteristics, which in turn introduces diverse ASR errors that need to be corrected through EC. 
While our training data inevitably contains some data similar to the benchmarking domains (e.g. daily conversation and presentations), we consider the overlap to be sufficiently small to regard all of them as OOD.\footnote{In fact, we confirmed that EC performs much better on in-domain data, i.e. unseen samples from the ASR system's training data, and keeps improving with more training steps.}


\begin{table*}[t!]
\centering
\begin{adjustbox}{max width=0.65\textwidth}
\setlength\tabcolsep{7pt}
\begin{tabular}{c|ccc}
\hline
Test & \qquad\qquad\qquad Domain \qquad\qquad\qquad & \# Utterances & Avg. Length \\
\hline
\hline
1 & business dialogue (spont.) & \phantom{0}187 & \phantom{0}9.34 \\
2 & university lecture (spont.) & \phantom{0}891 & 21.16 \\
3 & children stories (read) & 1649 & \phantom{0}9.03 \\
4 & proper nouns (read) & \phantom{0}350 & \phantom{0}3.86 \\
5 & agent-customer interactions (read)  & 1400 & \phantom{0}7.78 \\
6 & financial-domain dialogue (spont.) & \phantom{00}39 & 13.23 \\
7 & presentation (spont.) & 1046 & 16.08 \\
8 & presentation (spont.) & \phantom{0}448 & 12.89 \\
9 & miscellaneous (read) & 1600 & \phantom{0}8.04 \\
10 & daily conversation (read) & 1118 & \phantom{0}7.90 \\
11 & interview (spont.) & \phantom{0}378 & 25.79 \\
12 & presentation (spont.) & \phantom{0}490 & 23.48 \\
13 & customer support (read) & \phantom{0}332 & 13.34 \\
14 & \:\:financial-domain dialogue (spont.)\:\: & \phantom{0}291 & \phantom{0}9.97 \\
15 & daily conversation (spont.) & \phantom{0}586 & 10.56 \\
16 & addresses (read) & \phantom{0}150 & 10.71 \\
17 & miscellaneous (read) & 1050 & 10.39 \\
18 & miscellaneous (read) & \phantom{00}50 & 18.10 \\
19 & news (read) & \phantom{00}50 & 30.36 \\
20 & customer support (spont.) & \phantom{0}379 & 32.72 \\
21 & miscellaneous (spont.) & \phantom{0}500 & 12.40 \\
\hline
\end{tabular}
\end{adjustbox}
\caption{\label{tab:benchmark_info}
Benchmark details.
Our test sets encompass a wide range of domains, including both monologues/dialogues and spontaneous/read speech.
Average utterance lengths are computed with the Mecab tokenizer \citep{kudo2004applying}.
}
\end{table*}



\section{Experiments based on Sarashina-2 7B}
\label{sec:sarashina2_7b}

While Swallow-Mistral 7B is a continuously pretrained model built upon Mistral 7B \citep{jiang2023mistral}, Sarashina-2 7B is a recently opensourced Japanese LLM pretrained from scratch on a mixture of Japanese and English texts.
To verify that our conclusions are generalizable to different LLMs, we also run the whole experimental pipeline (\cref{sec:methods}-\cref{sec:experimental_results}) using Sarashina-2 7B.

\begin{figure}[t!]
    \includegraphics[width=0.95\columnwidth]{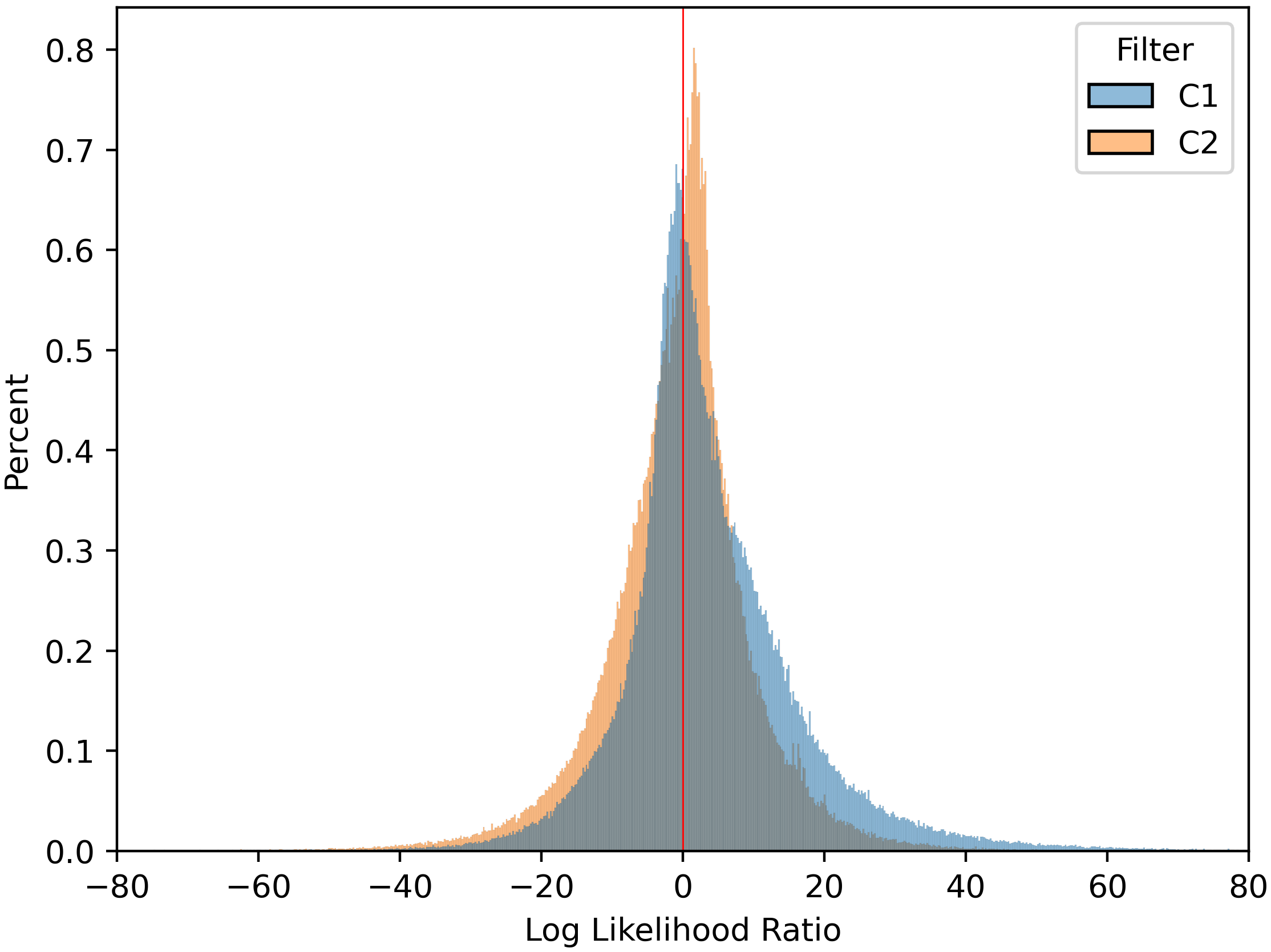}
    \caption{Log-likelihood ratio for the two criteria using Sarashina-2 7B. Red line shows the default threshold ($c_1 = c_2 = 1$).
    }
    \label{fig:log_likelihood_ratio_sarashina2}
\end{figure}

In Figure \ref{fig:log_likelihood_ratio_sarashina2}, we plot the distribution of the log-likelihood ratio for each criteria in our training data based on Sarashina-2 7B.
Out of the effective pairs (where ${W}^S \neq {W}^T$), 33\% are classified as noisy based on the C1 filter, 49\% based on C2 filter, and 63\% when combined.
While the C2 filter removes a larger portion of the data, we generally observe similar trends as Swallow-Mistral 7B.

In Table \ref{tab:sarashina2_results}, we report the results of our experiments using Sarashina-2 7B.
Similar to Swallow-Mistral 7B, we found that EC without data filtering fails to improve CER on average (11.84 $\rightarrow$ 11.84) due to overcorrection.
By applying our C1 filtering, we could significantly improve the average CER (11.84 $\rightarrow$ 11.41) whilst reducing the frequency of corrections.
Our C2 filtering has a similar benefit, and by combining both filters, we could significantly mitigate overcorrection and improve the original CER in nearly all (85.7\%; 18/21) of the test cases.
As in the case of Swallow-Mistral 7B, we found that inverse filtering generally has a negative effect on EC performance.

In Table \ref{tab:edit_results_sarashina2}, we show the results of edit distance based filtering using Sarashina-2 7B.
Again, we can confirm that simple filtering is much less effective compared to our sophisticated filtering which takes into account the pair-wise data quality and explicitly induces conservative behavior.

\bigskip

\bigskip

\begin{table}[t!]
\centering
\begin{adjustbox}{max width=0.95\columnwidth}
\setlength\tabcolsep{4pt}
\begin{tabular}{c||ccc|ccc}
\hline
\multirow{2}{*}{Test} & \multicolumn{3}{c|}{\underline{Edit Dist. (0.5)}} & \multicolumn{3}{c}{\underline{Edit Dist. (0.25)}}  \\
 & CER & \%EC & \%LA & CER & \%EC & \%LA  \\
\hline
\hline
Avg. & 11.87 & 40.6 & 58.8 & 12.01 & 39.7 & 62.5 \\
\hline
$<$Orig. & 61.9 & - & - & 47.6 & - & - \\
\hline
\hline
\end{tabular}
\end{adjustbox}
\caption{\label{tab:edit_results_sarashina2}
Results of EC using Sarashina-2 7B with data filtering based on maximum edit distance. 
}
\end{table}

\begin{table*}[t!]
\centering
\begin{adjustbox}{max width=0.99\textwidth}
\setlength\tabcolsep{4pt}
\begin{tabular}{c||c||ccc|ccc|ccc|ccc|ccc}
\hline
\multirow{2}{*}{Test} & \underline{Orig.} & \multicolumn{3}{c|}{\underline{No Filter}} & \multicolumn{3}{c|}{\underline{C1 Only}} & \multicolumn{3}{c|}{\underline{C2 Only}} & \multicolumn{3}{c|}{\underline{C1+C2}} & \multicolumn{3}{c}{\underline{Inv. C1+C2}} \\
 & CER & CER & \%EC & \%LA & CER & \%EC & \%LA & CER & \%EC & \%LA & CER & \%EC & \%LA & CER & \%EC & \%LA \\
\hline
\hline
1& \textbf{\phantom{0}6.66} & \phantom{0}9.25 & 33.9 & 62.7 & \phantom{0}7.97 & 13.8 & 58.3 & \phantom{0}7.87 & 14.4 & 72.0 & \phantom{0}7.30 & \phantom{0}6.3 & 54.5 & \phantom{0}7.60 & 10.3 & 55.6 \\
2& \phantom{0}8.18 & \phantom{0}8.09 & 54.4 & 58.9 & \textbf{\phantom{0}7.51} & 33.7 & 68.2 & \phantom{0}7.57 & 25.5 & 67.4 & \phantom{0}7.68 & 20.2 & 72.0 & \phantom{0}8.37 & \phantom{0}8.5 & 50.0 \\
3& 20.66 & 20.85 & 41.8 & 49.9 & \textbf{20.07} & 22.0 & 59.9 & 20.22 & 20.3 & 55.7 & 20.16 & 10.8 & 60.1 & 20.57 & \phantom{0}7.5 & 49.2 \\
4& 18.74 & \textbf{17.63} & 25.3 & 68.2 & 18.52 & 11.2 & 69.2 & 18.42 & \phantom{0}8.9 & 64.5 & 18.20 & \phantom{0}5.2 & 66.7 & 18.56 & \phantom{0}8.3 & 51.7 \\
5& \phantom{0}6.13 & \phantom{0}7.11 & 21.4 & 78.3 & \phantom{0}6.57 & \phantom{0}9.8 & 81.0 & \phantom{0}6.98 & 14.0 & 81.6 & \phantom{0}6.29 & \phantom{0}5.1 & 80.3 & \textbf{\phantom{0}5.91} & \phantom{0}3.9 & 50.9 \\
6& \phantom{0}7.20 & \textbf{\phantom{0}6.67} & 12.8 & 60.0 & \textbf{\phantom{0}6.67} & \phantom{0}7.7 & 66.7 & \phantom{0}6.99 & \phantom{0}7.7 & 33.3 & \phantom{0}6.89 & \phantom{0}5.1 & 50.0 & \phantom{0}6.99 & \phantom{0}2.6 & \!\!100.0 \\
7& 12.50 & 13.80 & 55.3 & 54.5 & 12.90 & 26.6 & 59.5 & 12.74 & 20.5 & 49.2 & \textbf{12.49} & 11.4 & 59.6 & 12.85 & 15.1 & 55.6 \\
8& \phantom{0}8.53 & \phantom{0}8.40 & 47.1 & 59.4 & \textbf{\phantom{0}7.89} & 25.0 & 64.7 & \phantom{0}8.27 & 23.8 & 59.8 & \phantom{0}8.10 & 12.8 & 69.2 & \phantom{0}8.45 & \phantom{0}9.8 & 47.5 \\
9& \phantom{0}8.47 & \phantom{0}7.92 & 33.6 & 52.1 & \textbf{\phantom{0}7.37} & 18.9 & 56.4 & \phantom{0}7.48 & 18.2 & 52.9 & \phantom{0}7.68 & 11.4 & 64.3 & \phantom{0}8.14 & 10.1 & 34.2 \\
10& \phantom{0}8.45 & \phantom{0}7.99 & 26.6 & 64.3 & \textbf{\phantom{0}7.78} & 14.7 & 67.7 & \phantom{0}7.81 & 12.0 & 70.9 & \phantom{0}7.85 & \phantom{0}7.1 & 72.2 & \phantom{0}8.35 & \phantom{0}4.7 & 39.6 \\
11& 19.77 & 19.59 & 50.1 & 58.8 & 19.64 & 24.8 & 58.9 & \textbf{19.30} & 18.2 & 59.1 & 19.70 & 12.4 & 60.0 & 21.74 & 11.6 & 59.5 \\
12& 12.02 & 11.94 & 46.0 & 56.4 & 11.65 & 24.1 & 60.2 & \textbf{11.55} & 20.4 & 52.0 & 12.01 & \phantom{0}9.6 & 48.9 & 11.95 & 11.4 & 50.0 \\
13& 13.06 & 12.99 & 27.5 & 60.5 & 13.01 & 12.5 & 69.2 & \textbf{12.86} & \phantom{0}8.0 & 56.0 & 12.92 & \phantom{0}3.8 & 33.3 & 13.00 & \phantom{0}4.8 & 66.7 \\
14& \textbf{26.10} & 28.14 & 46.8 & 63.3 & 26.44 & 16.7 & 66.7 & 26.70 & 13.3 & 61.3 & 26.19 & \phantom{0}7.3 & 52.9 & 26.81 & 12.9 & 70.0 \\
15& 15.23 & 16.12 & 45.9 & 53.5 & 15.28 & 21.2 & 54.9 & 15.32 & 16.9 & 45.6 & \textbf{15.14} & \phantom{0}9.0 & 56.2 & 15.45 & 12.9 & 46.4 \\
16& 12.03 & \phantom{0}9.06 & 41.3 & 77.4 & \textbf{\phantom{0}8.48} & 38.7 & 86.2 & \phantom{0}9.90 & 30.7 & 71.7 & 10.07 & 25.3 & 76.3 & 11.61 & \phantom{0}3.3 & \!\!100.0 \\
17& \phantom{0}9.98 & \phantom{0}9.49 & 26.5 & 63.0 & \phantom{0}9.42 & 17.2 & 71.3 & \textbf{\phantom{0}9.29} & 15.1 & 68.3 & \phantom{0}9.58 & 10.6 & 71.2 & \phantom{0}9.97 & \phantom{0}5.2 & 61.1 \\
18& 14.52 & 14.96 & 62.0 & 51.6 & 14.65 & 34.0 & 64.7 & 14.34 & 14.0 & 57.1 & \textbf{14.03} & \phantom{0}8.0 & \!\!100.0 & 14.65 & \phantom{0}2.0 & \!\!100.0 \\
19& \phantom{0}6.89 & \textbf{\phantom{0}5.16} & 58.0 & 79.3 & \phantom{0}5.46 & 40.0 & 85.0 & \phantom{0}6.10 & 44.0 & 86.4 & \phantom{0}5.91 & 30.0 & 86.7 & \phantom{0}6.78 & \phantom{0}6.0 & 66.7 \\
20& \phantom{0}7.81 & \phantom{0}7.38 & 66.7 & 69.4 & \textbf{\phantom{0}7.27} & 31.2 & 71.2 & \phantom{0}7.46 & 20.9 & 78.5 & \phantom{0}7.49 & 10.3 & 84.6 & \phantom{0}7.88 & \phantom{0}9.0 & 70.6 \\
21& \phantom{0}5.76 & \phantom{0}6.17 & 39.6 & 56.1 & \textbf{\phantom{0}4.98} & 21.0 & 70.5 & \phantom{0}5.26 & 17.6 & 62.5 & \phantom{0}5.19 & 12.4 & 71.0 & \phantom{0}5.60 & \phantom{0}5.8 & 48.3 \\
\hline
\hline
Avg. & 11.84 & 11.84 & 41.1 & 61.8 & \textbf{11.41} & 22.1 & 67.2 & 11.54 & 18.3 & 62.2 & 11.47 & 11.1 & 66.2 & 11.96 & \phantom{0}7.9 & 60.6 \\
\hline
$<$Orig. & - & 61.9 & - & - & 71.4 & - & - & 76.2 &- & -& \textbf{85.7} &- &- & 61.9 & -& -\\
\hline
\hline
\end{tabular}
\end{adjustbox}
\caption{\label{tab:sarashina2_results}
Results of EC using Sarashina-2 7B. Based on 21 internal test sets, we compute the macro average score for each metric (Avg.) and the ratio of test sets where CER is improved over the original ASR baseline ($<$Orig.).
}
\end{table*}

\end{document}